\pdfoutput=1

\documentclass[11pt]{article}

\usepackage{acl}

\usepackage{times}
\usepackage{latexsym}
\usepackage{verbatim}

\usepackage{subcaption}
\usepackage{graphicx}

\usepackage[T1]{fontenc}

\usepackage[utf8]{inputenc}

\usepackage{microtype}

\title{Do Data-based Curricula Work?}

\author{Maxim K. Surkov \unskip\enspace{\rm}\enspace Vladislav D. Mosin \unskip\enspace{\rm}\enspace Ivan P. Yamshchikov \\
  LEYA Lab, Yandex, Higher School of Economics \\}

\begin{document}
\maketitle
\begin{abstract}
Current state-of-the-art NLP systems use large neural networks that require extensive computational resources for training. Inspired by human knowledge acquisition, researchers have proposed curriculum learning - sequencing tasks (task-based curricula) or ordering and sampling the datasets (data-based curricula) that facilitate training. This work investigates the benefits of data-based curriculum learning for large language models such as BERT and T5. We experiment with various curricula based on complexity measures and different sampling strategies. Extensive experiments on several NLP tasks show that curricula based on various complexity measures rarely have any benefits, while random sampling performs either as well or better than curricula.
\end{abstract}

\section{Introduction}

In the last years state-of-art results in natural language processing (NLP) are often obtained with Transformer-like architectures based on the self-attention mechanism \cite{vaswani2017attention} such as BERT \citep{devlin-etal-2019-bert}, GPT-3 \cite{brown2020language}, T5 \cite{raffel2020exploring}, which could have billions of parameters. Due to many parameters, these architectures require lots of time and hardware resources to be trained.

Curriculum learning (CL) is one of the popular methods to reduce training time and increase the resulting quality of the model. Inspired by the importance of adequately ordering information when teaching humans \cite{avrahami1997teaching}, curriculum learning increases the difficulty of training samples shown to the model over time \cite{elman1993learning}. Previous studies have demonstrated that curriculum learning significantly impacts training time and quality in different machine learning domains, such as computer vision \cite{Soviany2020CurriculumLW} and reinforcement learning \cite{narvekar2020curriculum}. In NLP, some results hint that CL might be beneficial \cite{platanios-etal-2019-competence,xu2020curriculum,kocmi-bojar-2017-curriculum}; however, these results are not as optimistic as in reinforcement learning setup. 

We suggest dividing recent research in curriculum learning into two main categories: {\em task-driven} curriculum and {\em data-driven} curriculum. The idea of the task-driven curriculum was inspired by human behavior. First, the model learns how to solve a simple task, and then the difficulty is gradually increased. This type of curriculum proposed by \citet{bengio2009curriculum} is considered to be classical, and a majority of curriculum-related results are obtained in this framework. Alternatively to the task-driven curriculum, some curricula try to use some form of filtering or sorting of training data that could facilitate learning a model on a given task. We suggest calling these curricula {\em data-driven} and distinguishing them from the classical task-based approach.

This paper attempts to understand when data-driven curriculum learning works for transformer-based language models. Generally, data-driven curriculum learning is organized in two steps: first, estimating the complexity for the elements that comprise the dataset; second, designing a  sampling strategy, thus forming a curriculum. In the first part of the paper, we list potentially useful natural language processing complexity measures. The second part discusses possible sampling strategies that might apply to corresponding complexity measures. We run extensive experiments with different metrics and sampling strategies on three classes of NLP tasks: unsupervised learning with masked language modeling, text classification, and machine translation. Our experiments show that data-driven curriculum learning does not give quality increase or time reduction on all metric-sampling strategy setups and often makes results even worse. 

\begin{figure*}[h!]
	\centering
	\begin{subfigure}{.4\textwidth}
		\centering
    	\includegraphics[scale=0.22]{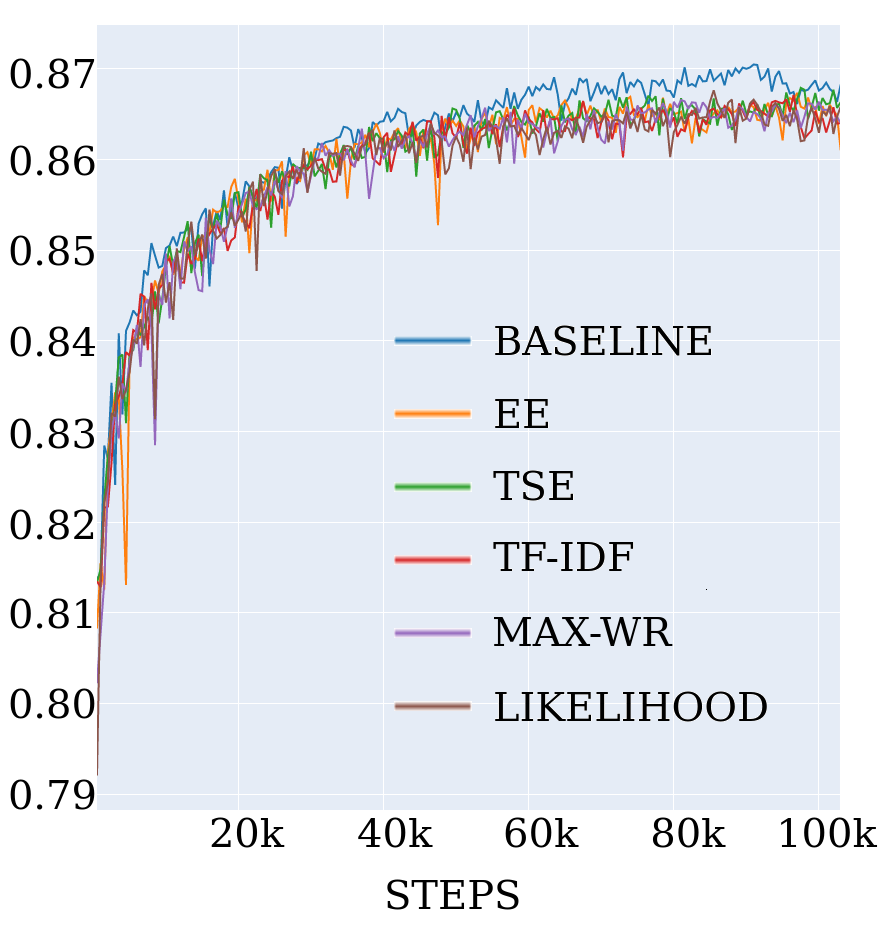}    
    	\caption{Sentiment140 with sort-merge sampler for all complexity measures.}
    	\label{fig:s140_SM_all_metrics}
	\end{subfigure}
	\begin{subfigure}{.4\textwidth}
		\centering
    	\includegraphics[scale=0.22]{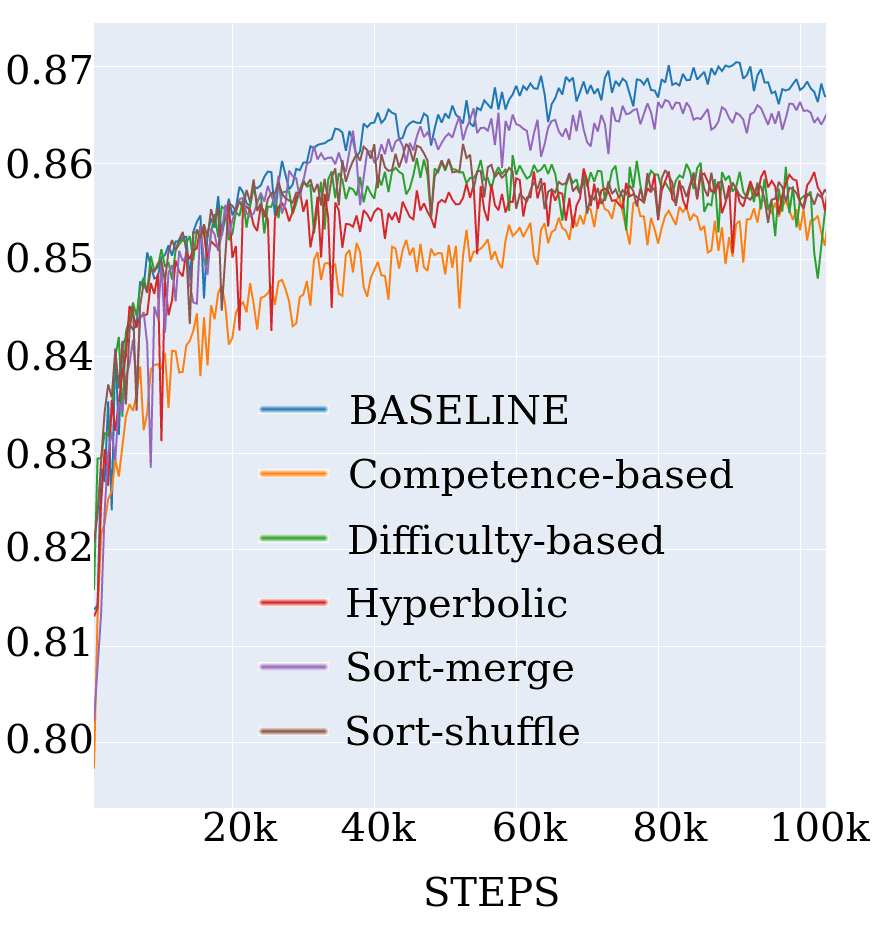}    
    	\caption{Sentiment140 with max word rank complexity measure for all samplers.}
    	\label{fig:s140_max_wf_rank_all_samplers}
	\end{subfigure}\\
	\begin{subfigure}{.4\textwidth}
		\centering
    	\includegraphics[scale=0.22]{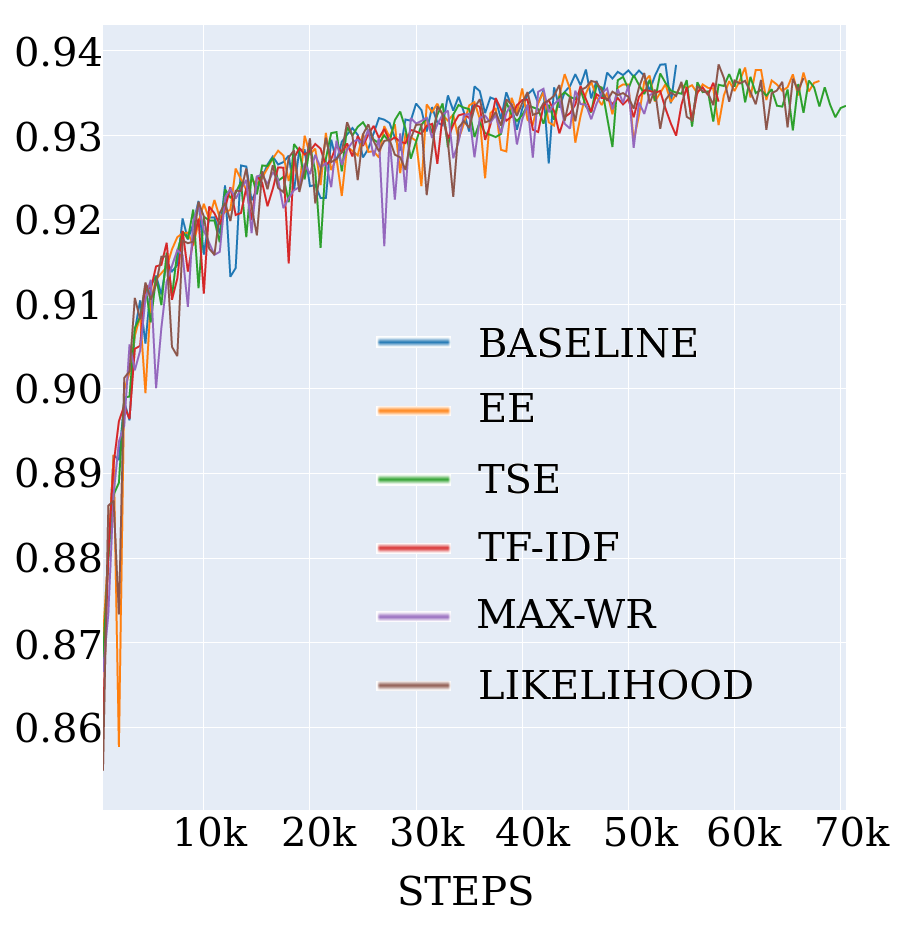}    
    	\caption{Hyperpartisan News with sort-shuffle samples for all complexity measures}
    	\label{fig:hnd_SS_all_metrics}
	\end{subfigure}
	\begin{subfigure}{.4\textwidth}
		\centering
        \includegraphics[scale=0.22]{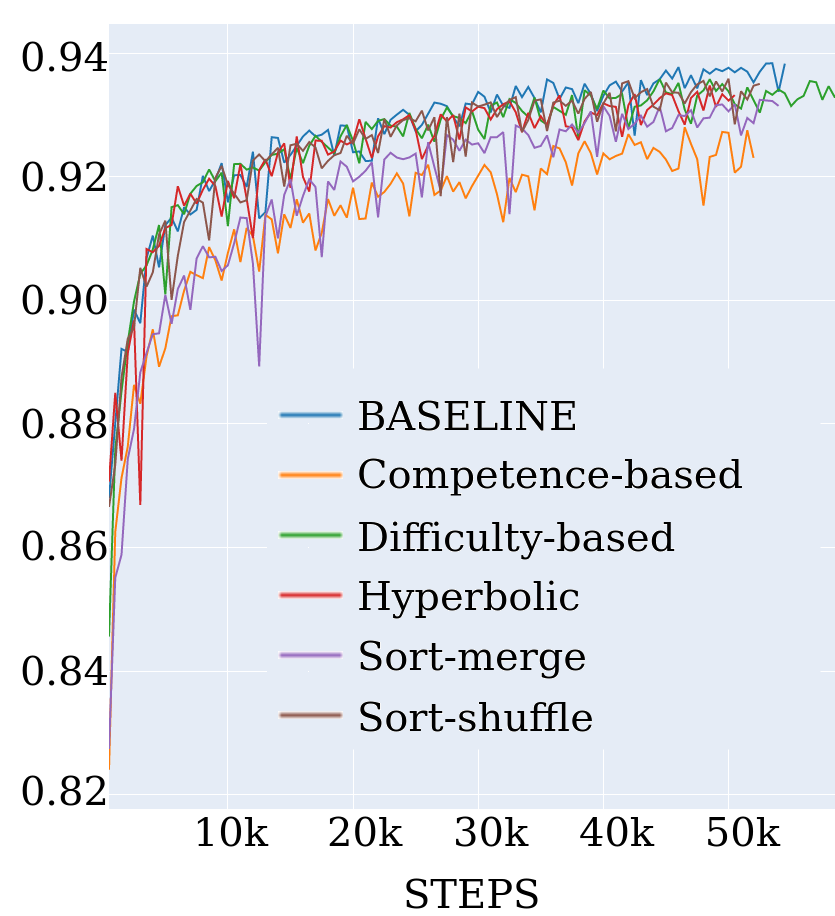}
        \caption{Hyperpartisan News with max word rank complexity measure for all samplers.}
        \label{fig:hnd_max_wf_rank_all_samplers}
	\end{subfigure}
	\caption{Pre-trained BERT fine-tuned on Sentiment140 and Hyperpartisan News Detection datasets. Accuracy of the classifier as a function of the number of training steps.}
	\label{fig:text_classification}
\end{figure*}

\section{Metrics}

The first important part of the curriculum learning pipeline is measuring the complexity of samples for a given dataset. Texts could have a complex structure, and one can measure their complexity in different ways. A variety of heuristically motivated methods is accompanied by several metrics based on specific aspects of information theory. For a review of heuristic text complexity measures such as length of TF-IDF \cite{aizawa2003information} we address the reader to Appendix \ref{a:heu}. In this paper, we also explore the metrics initially proposed by \citet{ay2006unifying} to measure the complexity of finite systems and try to see if one could apply these metrics to NLP tasks.

\citet{ay2006unifying} observes that for finite systems, a set of parts impacts the complexity of the system as well as inter-dependencies of the parts. In the context of NLP, this means that text is more than just a bag of words. The authors propose four different metrics to estimate the complexity of a system. However, one of these metrics maximizes on single-letter texts, such as "Aaaaaaaaa," while the second was created to measure cyclic sequences and does not apply to texts. Thus we experiment with two other metrics, namely, Tononi, Sporns, and Edelman (TSE) \cite{tononi1994measure} and excess entropy (EE), and adapt them to the complexity of texts. For the calculation of TSE and EE for NLP we address the reader to Appendix \ref{a:cal}.

\section{Samplers}

The second important part of curriculum learning is the sampling strategy (or sampler) - the algorithm deciding which samples should be shown to the model at which moment. Let us observe existing curricula and suggest some new ones. 

\textbf{Competence-based. CB}\\
A competence-based curriculum, offered by \citet{platanios-etal-2019-competence}, uniformly samples data from increasing dataset's prefix. Competence is a function $c(t)$, which defines the size of the dataset prefix.
$$c(t) = \min \left( 1, \sqrt{t\frac{1 - c_0^2}{T} + c_0^2}\right)$$
Where $T$ - total number of steps, $t$ - current step, $c_0$ - hyperparameter set to $0.01$.

\textbf{Hyperbolic. HYP} \\
The main idea of this sampler is to increase average batch complexity through time. All samples are split by complexity into $N$ sequential buckets with equal size. Training time is divided into $N$ epochs and the probability of sampling the element from the $j$-th bucket on the $i$-th epoch is proportional to the distance between $j$ and $i$.
$$Pr_i(j) = \frac{c}{|j - i|^{0.5}}$$
Where $Pr_i(j)$ - probability to sample from $j$-th bucket on the $i$-th epoch, $c$ - constant to guarantee that sum of all probabilities equals to $1$.

\textbf{Difficulty-based. DB} \\
This sampler is a reversed version of the competence-based one. A difficulty-based sampler takes elements from a linearly decreasing suffix instead of sampling from a gradually increasing prefix. 

\textbf{Sort-shuffle. SS}\\
All previously described samplers do not guarantee that the model would see each element in the training data. Sort-shuffle samples each element exactly once, randomly splitting the data into batches and sorting by average complexity.

\textbf{Sort-merge. SM}\\
Many complexity estimates correlate with the length of the text. The main idea of a sort-merge sampler is to remove this correlation and train the model on stable length distribution. This algorithm consists of four main steps: sort dataset by length; sequentially split into buckets; sort each bucket by a complexity metric; form $i$-th batch from $i$-th elements from each bucket. Like a sequential one, the sort-merge sampler shows each element to the model exactly once.

Equipped with the list of metrics and curriculum samplers, we can discuss our experimental results.

\section{Experiments}
We perform our experiments on three NLP tasks: text classification, machine translation (NMT), and masked language modeling (MLM). Here we discuss the first task of classification in detail. The extensive results of the experiments are available in Appendix \ref{a:ex}. All the experiments are performed with the HuggingFace library \cite{wolf-etal-2020-transformers}, which provides the models with their setups, such as hyperparameters and tokenizers. We did not change default parameters in our experiment unless specifically stated otherwise. Thus, the dataset and the model specify every experiment. We use the base version of the BERT model \cite{devlin-etal-2019-bert} for MLM and classification, and the small version of the T5 model \cite{raffel2020exploring} for machine translation. Experiments were performed on BooksCorpus\footnote{\url{https://huggingface.co/datasets/bookcorpus}} dataset for MLM, Sentiment140\footnote{\url{https://www.kaggle.com/kazanova/sentiment140}} and Hyperpartisan News Detection\footnote{\url{https://huggingface.co/datasets/hyperpartisan_news_detection}} for classification, and WMT16-en-de\footnote{\url{https://huggingface.co/datasets/wmt16}} for machine translation. To estimate the curriculum's convergence speed, we calculate the average number of steps to reach a threshold that is 10\% lower than the resulting saturation quality metric for every problem.

\subsection{Text Classification}

 Figure~\ref{fig:text_classification} summarizes the experiments with BERT for text classification. Neither different samplers nor complexity measures improve a BERT-based classifier's resulting accuracy. 
 
\subsection{Masked Language Modelling}

\begin{figure*}[t]
	\centering
	\begin{subfigure}{.25\textwidth}
		\centering
		\includegraphics[scale=0.2]{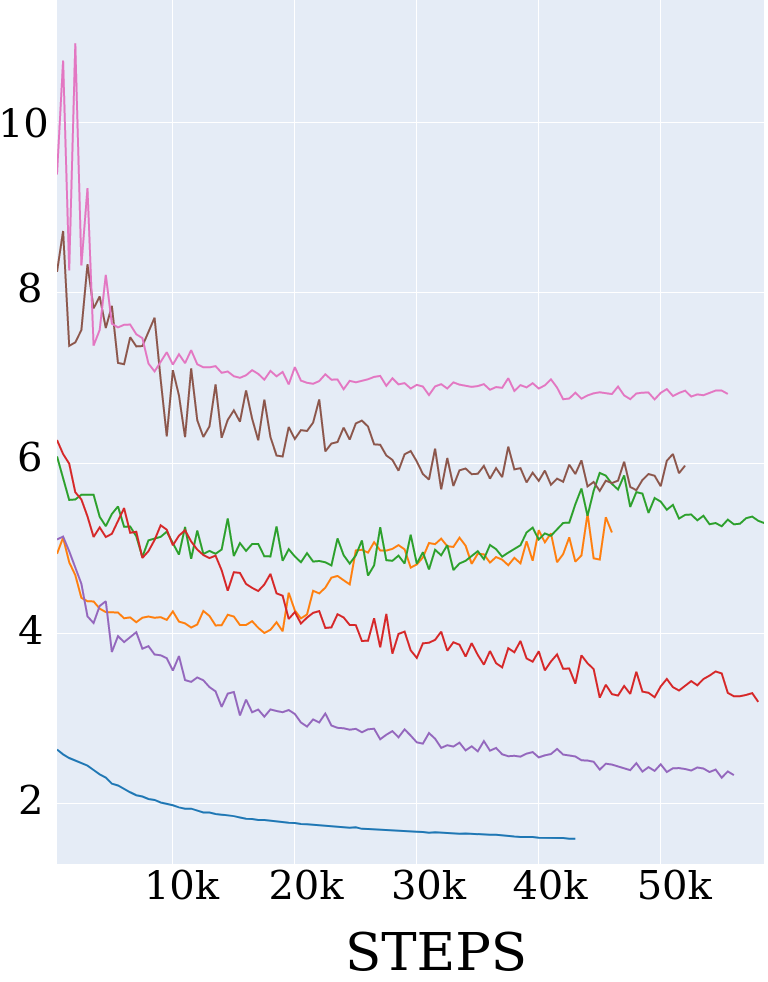}
		\caption{Competence-based}
		\label{fig:BooksCorpus_graphs_CB}
	\end{subfigure}
	\begin{subfigure}{.25\textwidth}
		\centering
		\includegraphics[scale=0.2]{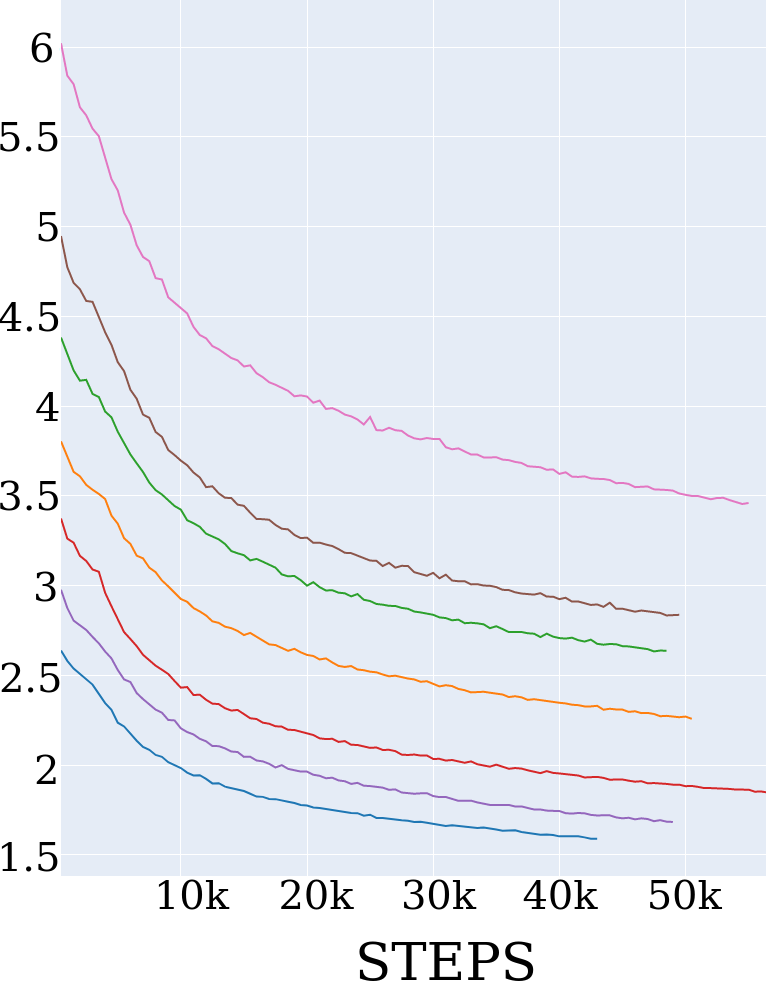}
		\caption{Difficulty-based}
		\label{fig:BooksCorpus_graphs_DB}
	\end{subfigure}
	\begin{subfigure}{0.25\textwidth}
		\centering
		\includegraphics[scale=0.2]{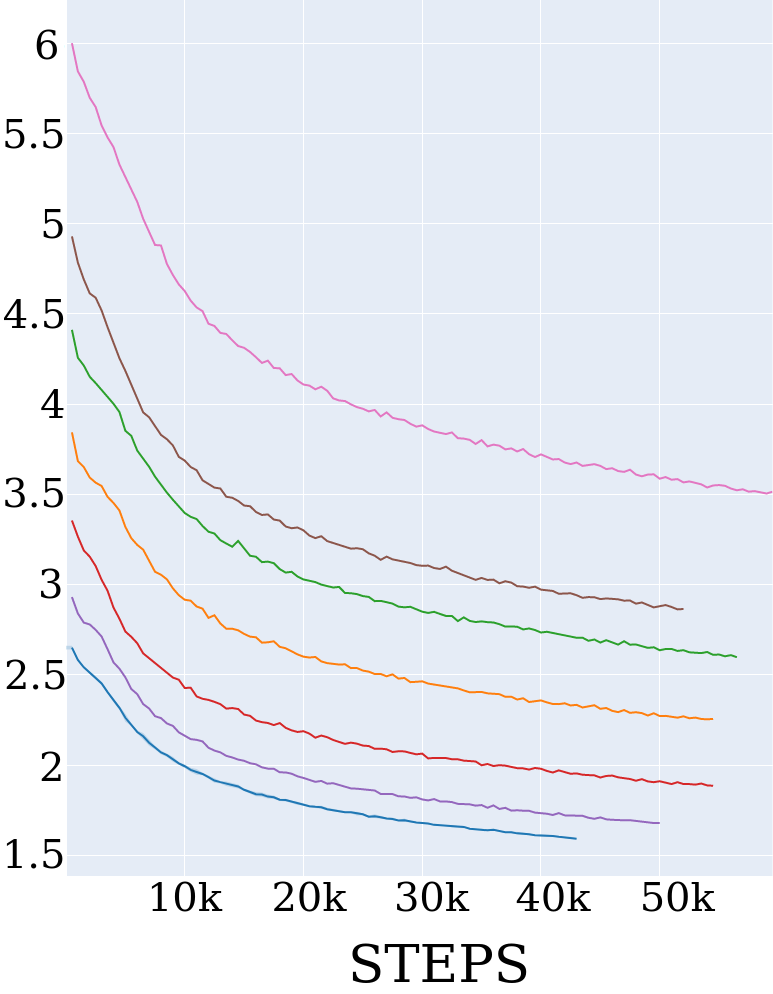}
		\caption{Hyperbolic}
		\label{fig:BooksCorpus_graphs_Hyp}
	\end{subfigure}\\
	\begin{subfigure}{.25\textwidth}
		\centering
		\includegraphics[scale=0.2]{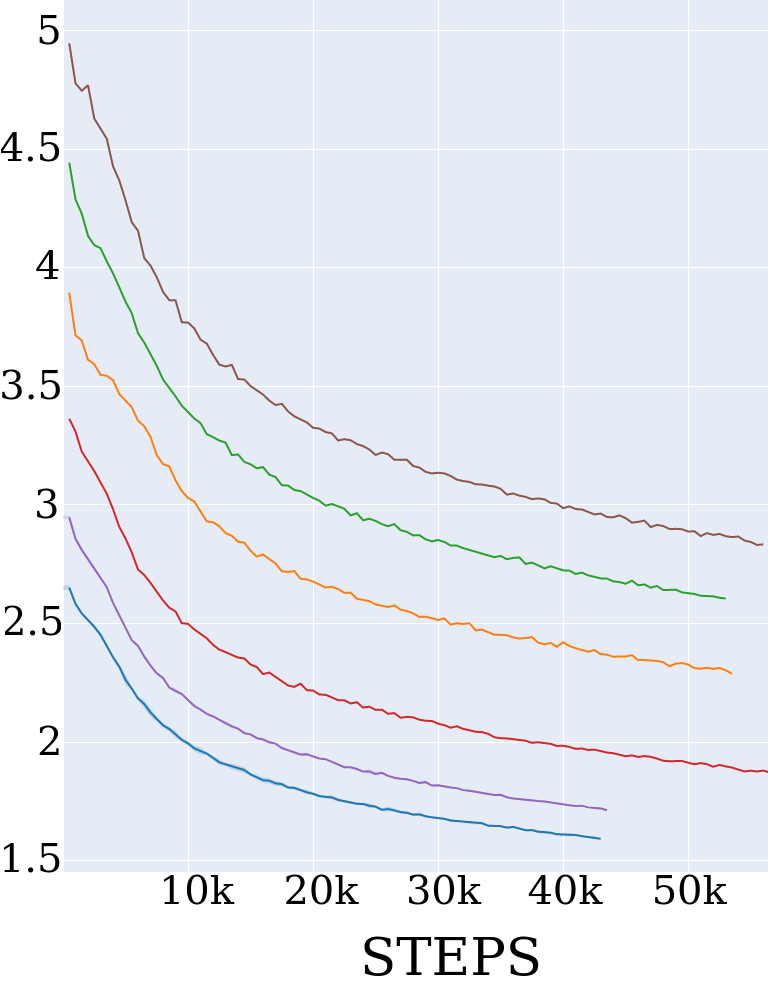}
		\caption{Sort-shuffle}
		\label{fig:BooksCorpus_graphs_SS}
	\end{subfigure}
	\begin{subfigure}{.25\textwidth}
		\centering
		\includegraphics[scale=0.2]{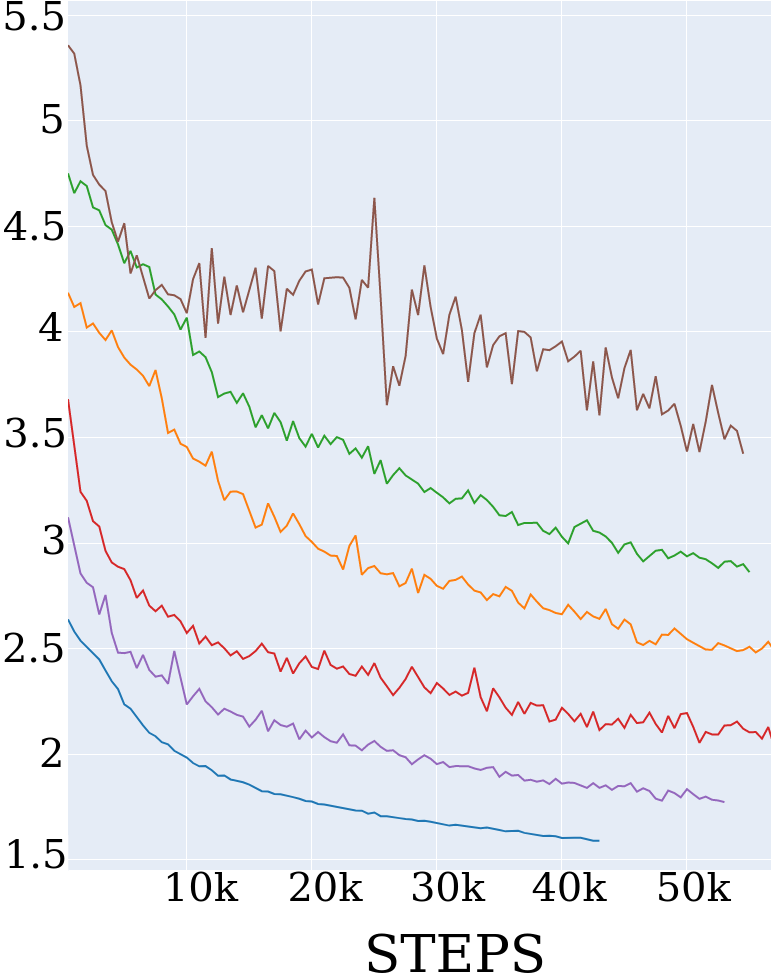}
		\caption{Sort-merge}
		\label{fig:BooksCorpus_graphs_SM}
	\end{subfigure}
	\begin{subfigure}{.25\textwidth}
		\centering
		\includegraphics[scale=0.5]{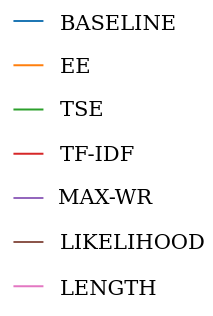}
		\label{fig:BooksCorpus_graphs_legend}
	\end{subfigure}
	\caption{Loss function dependency on the number of training steps on MLM for BooksCorpus dataset during the first 40k steps of training. Every plot depicts results for six different complexity estimates combined with a specific sampler.}
	\label{fig:BooksCorpus_graphs}
\end{figure*}

 Figure~\ref{fig:BooksCorpus_graphs} shows the results of MLM pretraining of BERT on BooksCorpus. Irrespective of sampling, the complexity measures have similar ranking in terms of their performance on MLM: length, likelihood, TSE, EE, TF-IDF, maximum word rank. Since sorted sampler takes length into account by design, it is not included in the corresponding plots. Data-based curricula show inferior results in comparison with the baseline.
 
\subsection{Neural Machine Translation}

Table~\ref{table:MT_quality} shows the experiments with T5 model \cite{raffel2020exploring} for machine translation and various curricula. We use the BLEU metric to estimate the quality of the resulting models. We calculate the average BLEU score over ten validations at saturation. Once again, curriculum learning does not give any notable benefits.

\begin{table}[h!]
    \caption{The average BLEU score from 50k to 100k steps on WMT16 dataset. Results better than the baseline are highlighted. '-' denotes the cases when complexity measure and sampler are not compatible.}
	\label{table:MT_quality}
	\centering
	\begin{tabular}{lcccccc}
		\hline
		Metrics & & \multicolumn{3}{c}{Samplers} &\\
		\hline
		& CB & DB & Hyp & SS & SM & \\
		\hline
		baseline & & \multicolumn{3}{c}{18.3} & \\
		\hline
		length &  10.1 & 17.4 & 16.3 & - & -\\
		TSE &  10.3 & \bf 18.4 &  16.8 &  13.8 &  14.8\\
		EE &  10.2 &  18.2 &  16.9 &  13.3 &  15.0\\
		\hline
	\end{tabular}
\end{table}

\section{Discussion}

We try to interpret obtained results cautiously. Though \citet{platanios-etal-2019-competence} report that competence-based sampling is beneficial for recurrent neural networks, we could not reproduce this result in transformer-based architectures. We also run experiments to check whether data-based curricula could work on non-transformer architectures. The results do not look encouraging; see Appendix \ref{a:plat}.

Curriculum learning depends on subtle factors, for example, a correct choice of hyperparameters. It is hard to check all possible values of hyperparameters, yet to the best of our capabilities, we address this issue in Appendix \ref{a:hyp}. The results do not seem to depend on the learning rate, and once again, curriculum learning shows no benefits.

At this point, we can only conclusively say two things: (1) a deeper investigation of the underlying information theoretic principles that stand behind curriculum learning is badly needed; (2) until we better understand these principles, data-based curriculum learning is a gamble with very low odds to gain either speed or resulting performance.

\section{Conclusion}
In this work, we ran extensive experiments with curriculum learning for transformer-based architectures on three NLP tasks: masked language modeling, text classification, and machine translation. We demonstrate that curricula do not help in the standard training setting and sometimes even worsen results.

\section{Acknowledgments}
The publication was supported by the grant for research centers in the field of AI provided by the Analytical Center for the Government of the Russian Federation (ACRF) in accordance with the agreement on the provision of subsidies (identifier of the agreement 000000D730321P5Q0002) and the agreement with HSE University  No. 70-2021-00139. This research was supported in part through computational resources of HPC facilities at HSE University \cite{kostenetskiy2021hpc}

\bibliography{anthology,custom}
\bibliographystyle{acl_natbib}

\appendix
\section{Heuristic Approaches to Text Complexity}
\label{a:heu}

The first idea is to determine the complexity of the text as its length. Despite its simplicity, this method is used in different works \cite{platanios-etal-2019-competence,kocmi-bojar-2017-curriculum}. The next family of approaches boils down to phonological, morphological, lexical, or syntactic metrics derived with some form of expert linguistic knowledge. However, \citet{van2010using} used Wikipedia and Simple Wikipedia corpora to demonstrate that language-based metrics do not correlate with the common sense text complexity. The third class of methods treats text as a bag of words and builds metrics based on the frequency analysis. For example, every word gets a rank equal to its position in the dictionary sorted by the number of word appearances in a corpus. In this case, complexity may be measured as a maximum rank among the words in a bag \cite{kocmi-bojar-2017-curriculum}. This metric is called max frequency rank. Another possible metric is called likelihood. The metric calculates the probability of the text under the assumption that all tokens are independent, just by multiplying probabilities of all tokens in the text \cite{platanios-etal-2019-competence}. Another metric from this group is TF-IDF \cite{aizawa2003information}, which is widely used in search systems. Finally, the last array of methods is based on using different neural network losses as a complexity measure of a sample.

\section{Using Information Theory for Text Complexity}
\label{a:cal}
 
Let $X_V = (X_{v1}, X_{v2}, \ldots)$ be a sequence of random variables from set $V = (v1, v2, \ldots)$, and $A$ is a subset of $V$, then $X_A$ is a subsequence of $X_V$ with elements from $A$. Let's determine $H(X_A)$ as entropy of sequence $X_A$. However, texts consist of words or tokens, not random variables. We propose the following procedure of transforming texts into random variable sequences. For each token in position $i$ we compute the percentage of texts with this token on the same position and replace the original token with binary distribution with a probability of one equal to the calculated percentage. After transforming text into a sequence of random variables, we can compute its entropy. 
\begin{eqnarray}
H(X_V) &=& H(X_{v1}) + H(X_{v2} | X_{v1}) \nonumber \\
&+& H(X_{v3} | X_{v2}, X_{v1}) + \ldots \nonumber
\end{eqnarray}

If one wants to apply this formula, one must compute entropy for many different conditional distributions while these distributions depend on the order of tokens in a text. First, direct application of the formula would overfit a specific text since all texts are different in a corpus. Second, such computation could not be carried out in a reasonable time. The limit context for conditional distributions to the nearest neighbors one obtains the following formula
$$H(X_V) = H(X_{v1}) + \sum\limits_{i = 2}^{\#V} H(X_{v_i} | X_{v_{i - 1}})$$

Using this approximation for entropy one can compute  excess entropy (EE) and the complexity measure
Tononi, Sporns and Edelman (TSE), \cite{tononi1994measure} as they are formulated by \citet{ay2006unifying}
\begin{equation}\label{eq:ee}
    EE(X_V) = \left [ \sum\limits_{v \in V} H(X_{V \textbackslash v}) \right ] - (n - 1)H(X_V),
\end{equation}
\begin{equation}\label{eq:tse}
TSE(X_V) = \sum\limits_{k = 1}^{n - 1}\frac{k}{n}C^{(k)}(X_V),
\end{equation}
where $n$ is a size of set $V$ and
$$C^{(k)}(X_V) = \frac{n}{k {n \choose k}}\sum\limits_{A \subseteq V, |A|=k} H(X_A) - H(X_V).$$

\section{Additional Experiments}
\label{a:ex}

\subsection{Convergence Speed}

Curriculum learning is often apprised for the speed-up of the model's convergence. The intuition here is to provide a curriculum that would help to achieve the same result faster, yet without a significant loss in quality. We carried out several experiments to see if data-based curricula could speed up the learning in transformer-based language models.

\subsubsection{Classification}

Tables \ref{table:hnd_fine_tuning} \ref{table:s140_fine_tuning} show average number of training steps needed to reach 90\% of the resulting accuracy for the corresponding classification task. On Sentiment140 TF-IDF, TSE, and maximum word rank speed the convergence up to 3\% with some samplers. However, other metrics or sampling strategies slow down the model's convergence speed, while on a bigger HND dataset, other curricula show results better than the baseline. One could conclusively say that length is the worse metric to organize curriculum in all experiment configurations. The one more important conclusion is that the model can not always estimate the complexity of the sample concerning its' internal state (MLM-loss does not speed up the training speed and drawdown the final model quality on the Sentiment140 dataset). This happens when the model is expressive enough, and all samples have equal complexity in model-based metrics.

\begin{table*}[ht]
    \caption{The average number of steps needed to reach given threshold for all configurations metric-sampler on text classification task on Hyperpartisan News Detections dataset. Maximal deviation for 3 runs is less than $3k$ steps. Results better than the baseline are highlighted. $\infty$ means that model did not reach the threshold, '-' denotes the cases when complexity measure and sampler are not compatible.}
	\label{table:hnd_fine_tuning}
	\centering
	\begin{tabular}{lccccccc}
		\hline
		Metrics & Threshold& Accuracy  & & \multicolumn{3}{c}{Samplers} & \\
		& & & CB & DB & Hyp & SS & SM \\
		\hline
		baseline & 92.9\%&  93.8\%   & & \multicolumn{3}{c}{22k} & \\
		\hline
		length & 92.9\%& 93.7\% & 55k & 23k & 22.5k & - & -  \\
		TF-IDF & 92.9\%& 93.5\%  & $\infty$ & \bf 19.5k & 24k & 23.5k & 33k  \\
		TSE & 92.9\%& 93.8\% & 56.5k &  21k & 23k & 22k & 31k  \\
		EE & 92.9\%& 93.8\%  & 71.5k & 25.5k & 22.5k & {\bf 19.5k} & 32.5k  \\
		max wr & 92.9\%& 93.6\%  & $\infty$ & 22k & \bf 20.5k & 22.5k & 39k  \\
		likelihood & 92.9\% & 93.8\% & $\infty$ & \bf 20k & 24k & \bf 20k & 30k \\
		MLM-loss & 92.9\% & \bf 93.9\% & 23.5k & {\bf 18k} & 23k & 24k & \bf 20k \\
		\hline
	\end{tabular}
\end{table*}

\begin{table*}[ht]
    \caption{The average number of steps needed to reach given threshold for all configurations metric-sampler on text classification task on sentiment140 dataset. Maximal deviation for 3 runs is less than $3k$ steps. Results better than the baseline are highlighted. $\infty$ means that model did not reach the threshold, '-' denotes the cases when complexity measure and sampler are not compatible.}
	\label{table:s140_fine_tuning}
	\centering
	\begin{tabular}{lccccccc}
		\hline
		Metrics & Threshold & Accuracy & & \multicolumn{3}{c}{Samplers} &\\

		& & & CB & DB & Hyp & SS & SM \\
		\hline
		baseline & 85.5\%& {\bf 87\%} & & \multicolumn{3}{c}{17.5k} &  \\
		\hline
		length & 85.5\% &  86.2\%&  112.5k & 20k & 19k & - & - \\
		TF-IDF & 85.5\%& 86.7\% & 115.5k & 21.5k & 19.5k &  \bf 16.5k & 22k  \\
		TSE  & 85.5\% & 86.8\% & 95.5k & \bf 16.5k & 20.5k & 21.5k & 18k  \\
		EE & 85.5\% & 86.7\%  & 59k & 19.3k & 23k & 20k & 19k \\
		max wr & 85.5\% & 86.7\% & 70k & 18.5k & 19.5k & \bf 17k & 19k \\
		likelihood & 85.5\% & 86.7\% & 112k & 17.5k & 21.5k & 17.5k & 21.5k  \\
		MLM-loss & 85.5\%& 86.1\% & 59.5k & 21k & 23.5k & 19.5k & 20k \\
		\hline

	\end{tabular}
\end{table*}

\subsubsection{Pretraining MLM}

 Figure~\ref{fig:BooksCorpus_graphs} shows a significant slowdown in model convergence speed can be seen for all curricula compared to the baseline learning regime. One can also divide all metrics into two distinct groups. The first one consists of maximum word rank and TF-IDF. The second group includes EE, TSE, likelihood, and length. The metrics in the first group allow the model to converge to a lower loss value. However, the second group's metrics hinder the convergence and seem to have higher saturation loss. Hence, it isn't easy to find a universal threshold to reasonably compare all metrics and samplers. One should also note that only maximum word rank does not degrade the model quality compared to the baseline, while other curricula cause severe deterioration. Finally, the last main observation is that curriculum learning, unfortunately, does not allow us to run MLM faster. Moreover, the number of training steps needed to reach a given threshold could be several times higher in comparison with the baseline approach. Table~\ref{table:BooksCorpus_pretraining} illustrates this fact.

\begin{table*}[ht]
    \caption{The average number of steps needed to reach given threshold for all configurations metric-sampler on pretraining on BooksCorpus dataset. Maximal deviation for 3 runs is less than $3k$ steps. All complexity measures based curricula reach saturation at higher losses than the baseline thus we used an arbitrary threshold of 3.5 for them. Results better than the baseline are highlighted. $\infty$ means that model did not reach the threshold, '-' denotes the cases when complexity measure and sampler are not compatible.}
    \label{table:BooksCorpus_pretraining}
	\centering
	\begin{tabular}{lccccccc}
		\hline
		Metrics & Threshold & Saturation & & \multicolumn{3}{c}{Samplers} & \\

		& Loss & Loss  & CB & DB & Hyp & SS & SM  \\
			\hline
		baseline & 2.00 &  {\bf 1.58} & & \multicolumn{3}{c}{ 9.5k} &\\
		\hline
		\hline
		max wr & 2.00 & {\bf 1.58} & $\infty$ & 17.5k & 16.5k & 16.5k & 27k \\
		TF-IDF & 2.00 & 1.84 & $\infty$ & 34k & 35k & 37.5k & $\infty$ \\
		\hline
		\hline
		EE & 3.50 & 2.25 & $\infty$ & 4k & \bf 3.5k & 4.5k & 9.5k  \\
		TSE & 3.50 &  2.60 & $\infty$ & 9k & 9k & 8.5k & 18k \\
		likelihood & 3.50 & 2.83 & $\infty$ & 13.5k & 13.5k & 15.5k & 50k \\
		length & 3.50 & 3.45 & $\infty$ & 50.5k & $\infty$ & - & - \\
		\hline

	\end{tabular}
\end{table*}

\subsection{Data-based Curricula for Other Architectures}
\label{a:plat}

\begin{figure*}[t!]
	\centering
	\begin{subfigure}{.4\textwidth}
		\centering
    	\includegraphics[scale=0.5]{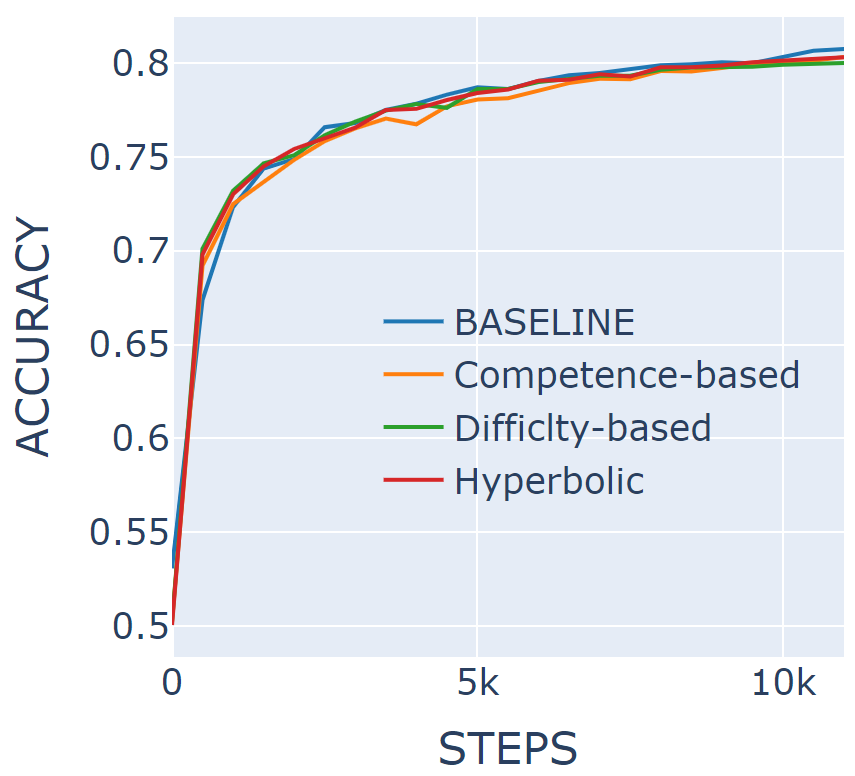}    
    	\caption{Sentiment140 with length as complexity metric and three samplers.}
    	\label{fig:lstm_len}
	\end{subfigure}
	\begin{subfigure}{.4\textwidth}
		\centering
    	\includegraphics[scale=0.5]{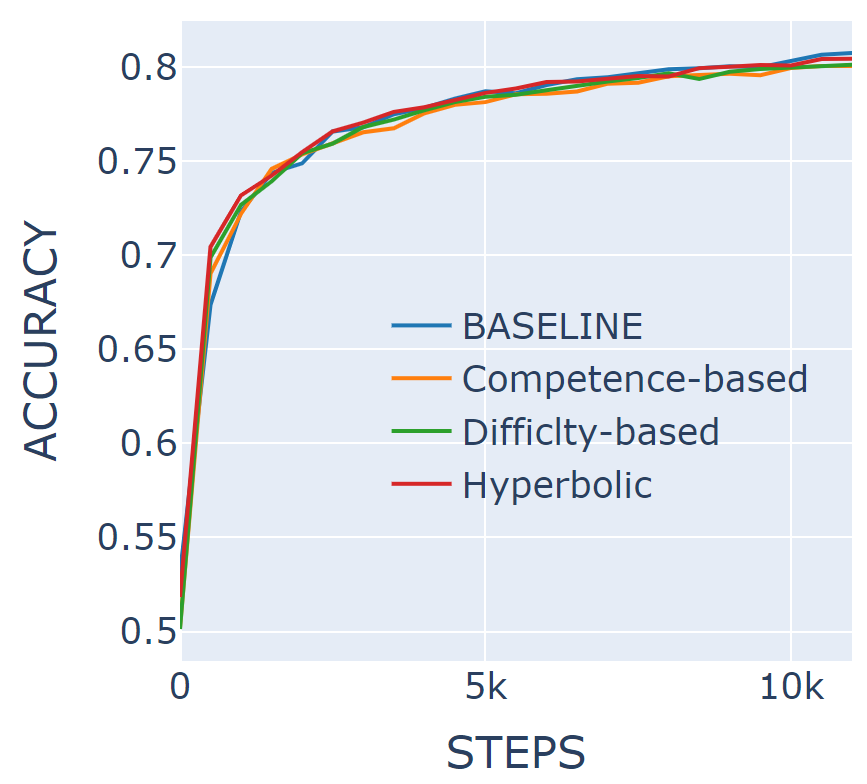}    
    	\caption{Sentiment140 with TSE as complexity metric and three samplers.}
    	\label{fig:lstm_tse}
	\end{subfigure}\\
	\caption{Test results with LSTM on Sentiment140 dataset. Accuracy of the classifier as a function of the number of training steps.}
	\label{fig:text_classification_lstm}
\end{figure*}

It seems that data-based curriculum learning cannot increase quality or reduce training time for transformer-based models. Though \citet{platanios-etal-2019-competence} report that competence-based sampling is beneficial for recurrent neural networks, we could not reproduce this result in transformer-based architectures. While some curricula might be useful for smaller architectures on some tasks, they have no significant benefits for larger architectures. Let us double-check that with the recurrent neural network architecture to see if the negative result obtained above is associated with certain properties of attention-based architectures or could be reproduced with various artificial neural networks. We run our experiments on Sentiment 140 with $90\%$ train and $10\%$ test split. The curricula include Hyperbole, Difficulty-Based and Competence-Based samplers, and TSE and length difficulty metrics. Figure~\ref{fig:text_classification_lstm} shows that data-driven curricula do not have a significant influence on the results.

Comparing Figure~\ref{fig:text_classification_lstm} with Tables \ref{table:s140_fine_tuning} -- \ref{table:hnd_fine_tuning} one could see that data-based curricula are hardly beneficial even for smaller architectures. Rather, under certain conditions, one could get some improvement of convergence, yet on a different task, the same choice of complexity measure and sampling strategy would be on par with the baseline.

\subsection{Data-based curricula and Hyperparameters}
\label{a:hyp}

\begin{figure*}[t]
	\centering
	\begin{subfigure}{.3\textwidth}
		\centering
    	\includegraphics[scale=0.32]{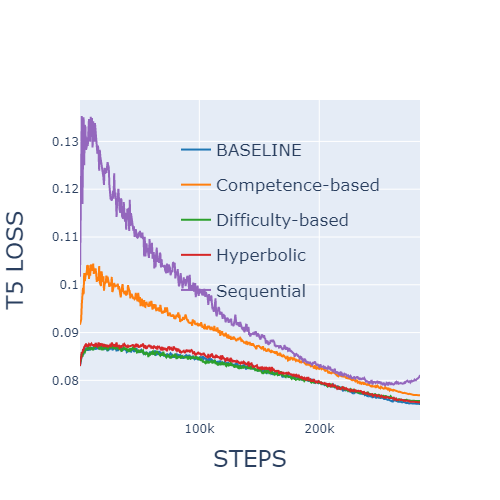}    
    	\caption{learning rate $10^{-3}$}
    	\label{fig:len_lr001}
	\end{subfigure}
	\begin{subfigure}{.3\textwidth}
		\centering
    	\includegraphics[scale=0.32]{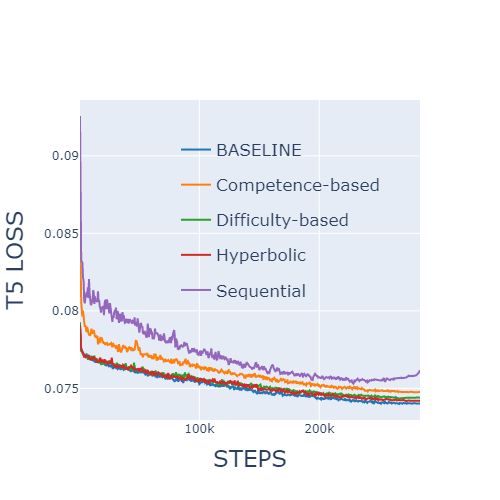}    
    	\caption{learning rate $10^{-4}$}
    	\label{fig:len_lr0001}
	\end{subfigure}
	\begin{subfigure}{.3\textwidth}
		\centering
    	\includegraphics[scale=0.32]{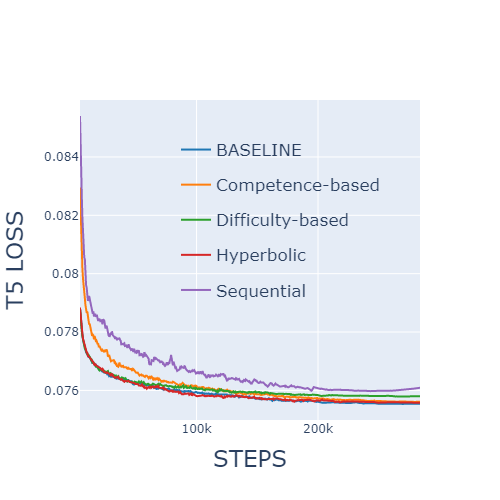}    
    	\caption{learning rate $10^{-3}$}
    	\label{fig:len_lr00001}
	\end{subfigure}\\
	\caption{Test results for NMT on WMT16 with different learning rates with length complexity measure}
	\label{fig:len_lr}
\end{figure*}

Extensive experiments on different NLP tasks show that data-based curriculum learning does not help to increase quality with default hyperparameters. Hyperparameters' importance for the curriculum is an open question. Some papers state that hyperparameters, especially learning rate, are essential for curriculum \cite{zhang2018empirical}. On the other hand, some papers propose methods that are not highly sensitive to hyperparameters \cite{platanios-etal-2019-competence}. It seems that hyperparameters choice is discussed mainly in the works addressing NMT, so we run additional experiments with our curricula and three different learning rates ($10^{-3}$, $10^{-4}$, $10^{-5}$) on NMT as well. Results demonstrate that models' behavior does not depend on the learning rate much, and for every learning rate, curricula do not give a significant quality increase. Results for excess entropy are presented in Figure \ref{fig:len_lr}.

\begin{figure*}[t]
	\centering
	\begin{subfigure}{.3\textwidth}
		\centering
    	\includegraphics[scale=0.32]{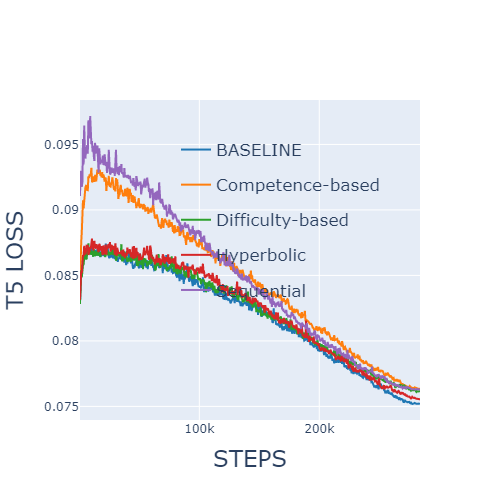}    
    	\caption{learning rate $10^{-3}$}
    	\label{fig:ee_lr001}
	\end{subfigure}
	\begin{subfigure}{.3\textwidth}
		\centering
    	\includegraphics[scale=0.32]{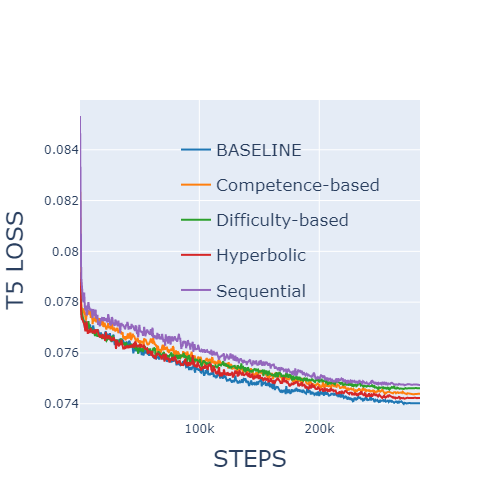}    
    	\caption{learning rate $10^{-4}$}
    	\label{fig:ee_lr0001}
	\end{subfigure}
	\begin{subfigure}{.3\textwidth}
		\centering
    	\includegraphics[scale=0.32]{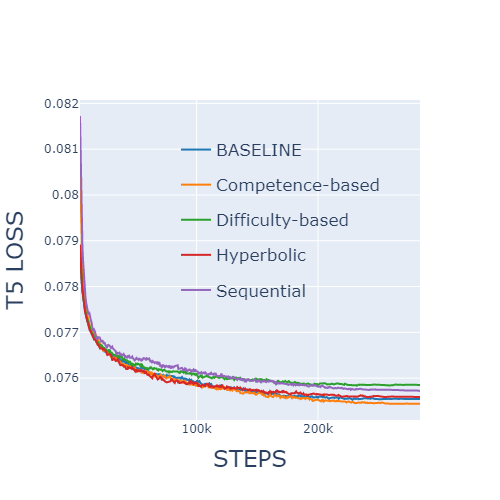}    
    	\caption{learning rate $10^{-5}$}
    	\label{fig:ee_lr00001}
	\end{subfigure}\\
	\caption{Test results for NMT on WMT16 with different learning rates with excess entropy as a complexity measure}
	\label{fig:ee_lr}
\end{figure*}

\begin{figure*}[t]
	\begin{subfigure}{.3\textwidth}
		\centering
    	\includegraphics[scale=0.32]{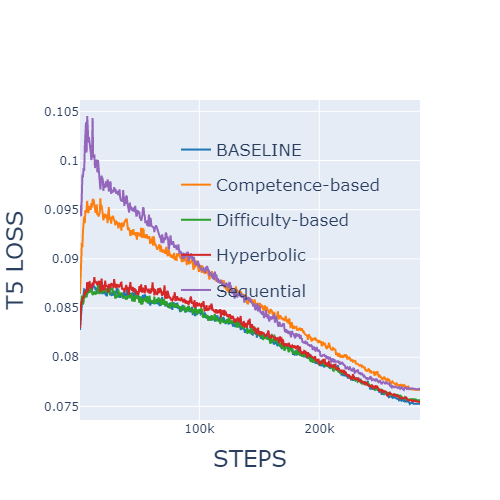}    
    	\caption{learning rate $10^{-3}$}
    	\label{fig:tse_lr001}
	\end{subfigure}
	\begin{subfigure}{.3\textwidth}
		\centering
    	\includegraphics[scale=0.32]{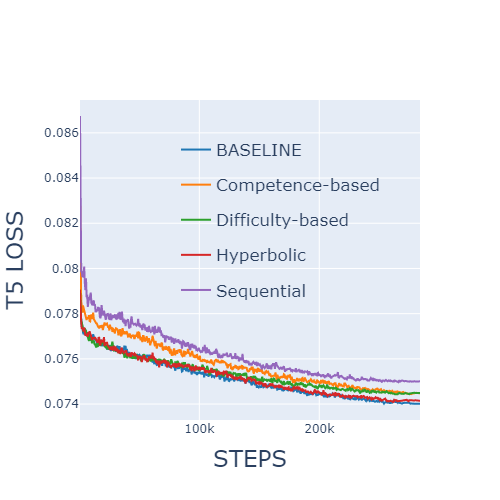}    
    	\caption{learning rate $10^{-4}$}
    	\label{fig:tse_lr0001}
	\end{subfigure}
	\begin{subfigure}{.3\textwidth}
		\centering
    	\includegraphics[scale=0.32]{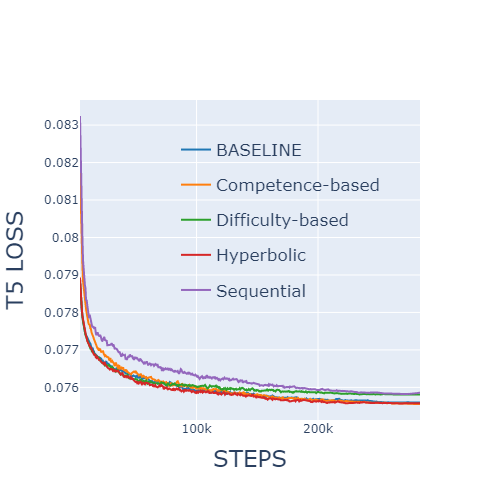}    
    	\caption{learning rate $10^{-5}$}
    	\label{fig:tse_lr00001}
	\end{subfigure}\\
	\caption{Test results for NMT on WMT16 with different learning rates with TSE as a complexity measure}
	\label{fig:tse_lr}
\end{figure*}

\end{document}